\documentclass[11pt]{article}

\usepackage[preprint]{acl}

\usepackage{times}
\usepackage{latexsym}

\usepackage[T1]{fontenc}
\usepackage[utf8]{inputenc}

\usepackage{microtype}

\usepackage{inconsolata}

\usepackage{graphicx}
\usepackage{placeins}
\usepackage{xspace}
\usepackage{xcolor}
\usepackage{tikz}
 \usepackage{booktabs}
\usepackage{amsfonts} 
 \usepackage{array} 

\newcommand{\method}{\textsc{APD}\xspace}
\newtheorem{proposition}{Proposition}

\title{Structural Automata for Neural Semantic Parsing}

\author{Xiao Zhang \\ University of Groningen \\ xiao\_is\_npc@outlook.com}

\begin{document}
\maketitle

\begin{abstract}
Neural semantic parsers generate meaning representations token by token, yet
well-formedness depends on references, scope, and graph relations that unfold
during generation. Constrained decoding prevents locally invalid tokens but
does not guarantee that a valid prefix can be completed within a fixed
output-length limit. We investigate structural automata at three stages of
DRS parsing (decoding, post hoc processing, and training) and additionally
evaluate decoding on AMR. For decoding, we introduce accepting-path decoding
(APD), which maintains a verified completion within the remaining length limit
and follows the model's preferred token whenever that token can also be
completed. APD guarantees an accepted output without retracting tokens. Under
tight length limits, it improves Smatch++ over local masking by 5.1 and 2.4
points on DRS and by 6.4 and 3.2 points on AMR, comparable to rollback under
the same limit, which retracts tokens. For post hoc processing,
invalid outputs concentrate on the same inputs: eight unconstrained samples
yield no valid output for 10.2\% of inputs with limited training data,
compared with 0.4\% with full data, where filtering comes within 0.2 Smatch++
points of APD. For training, penalizing automaton-rejected tokens improves
development Smatch++ by 3.6--11.0 points after 10\% of training, but this
advantage diminishes as cross-entropy alone reduces the probability assigned
to rejected tokens from 14\% to below 2\% within the first sixth of training with limited data.
\end{abstract}

\section{Introduction}

Neural semantic parsers generate meaning representations token by token, but
their token probabilities do not explicitly enforce the structural
requirements of well-formed outputs. Symbolic constraints make these
requirements explicit and have been used to guide learning
\citep{english-etal-2025-grammar,ahmed-etal-2023-pseudo}, restrict decoding
\citep{scholak-etal-2021-picard,geng-etal-2023-grammar}, and validate or
repair generated outputs \citep{liu-2024-soft,ugare-etal-2025-itergen}. These approaches act at different stages and target different failure modes, which raises a question: when and how should
structural constraints act in neural semantic parsing?

Meaning representations such as AMR \citep{banarescu-etal-2013-abstract},
Discourse Representation Structures (DRS;
\citealp{abzianidze-etal-2017-parallel}), and executable queries
\citep{scholak-etal-2021-picard} impose structural dependencies that extend
beyond individual tokens: references must resolve, variables must be bound,
and scope constraints must be respected. In DRS, a forward reference such as \texttt{Patient +2} requires a
target concept to be introduced later; a proposition requires a new box to be
opened; and discourse referents must remain accessible. Such requirements may remain unresolved across many decoding steps, so
local token validity is insufficient in two ways. Under a fixed maximum
output length (the output budget), a valid token may lead to a prefix that
cannot be completed within the remaining budget. Moreover, a sequence
accepted by a context-free grammar may still violate graph-level constraints.
A structural automaton makes these requirements executable: it tracks the evolving state, validates completed outputs, and checks whether a partial output can still be completed.

We use a single structural automaton at three stages of DRS parsing, each raising a distinct question. During decoding, the automaton constrains token choices, and the question is whether the prefix can still be completed within the budget. After generation, it validates and repairs outputs, and the
question is whether every input receives a valid output. During training, it
identifies structurally invalid next tokens, and the question is how much
probability the model still assigns to them. These questions yield three
measurable quantities: the distance to acceptance, the per-input failure
tail, and the rejected-token mass.

For decoding, we introduce accepting-path decoding (APD), which maintains a
verified completion of the current prefix within the remaining budget. At
each step, APD follows the model's preferred token whenever that token also
admits a verified completion; otherwise, it advances along the previously
verified path. APD therefore always produces an automaton-accepted output
within the budget, without retracting emitted tokens, even when the search
for a new completion fails. APD requires only state transitions, an
acceptance test, and a completion procedure, so it extends to other representations, as our AMR results confirm.

Our experiments address these questions in turn. Under two tight output
budgets, APD improves Smatch++ over local masking by 5.05 and 2.44 points on
DRS and by 6.44 and 3.20 points on AMR, by completing graphs that local masking leaves truncated. Same-budget rollback achieves comparable quality but
retracts up to 27 emitted tokens. When we withhold a fraction of new
completions during decoding, a decoder that recomputes viability at every
step fails on 29--70\% of DRS inputs, whereas APD completes every input by
following its verified completion. For post hoc processing, invalid outputs
concentrate on a persistent subset of inputs: with a 600-sentence training
subset, eight unconstrained samples leave 10.2\% of inputs without a valid
output, 13 times the share expected if samples failed independently. For
training, penalizing automaton-rejected tokens accelerates early
optimization, but the advantage diminishes as cross-entropy alone reduces the
rejected-token mass below 2\% within the first sixth of training. Our
contributions are:
\begin{itemize}
\item \textbf{An incremental structural automaton for DRS} that operates
directly on subword tokens and supports state transitions, acceptance
checking, and verified completion.
\item \textbf{Accepting-path decoding (APD)}, which guarantees an
automaton-accepted output within a fixed length budget without retracting
emitted tokens, evaluated on DRS and AMR against local masking, per-step viability, and same-budget rollback in a matched study.

\item \textbf{A unified analysis of the three stages} through the distance to acceptance, the per-input failure tail, and the rejected-token mass, showing which quantity limits each stage.
\end{itemize}

\section{Related Work}
\label{sec:related}

\paragraph{Constraints during decoding.}
Constrained decoding restricts the tokens available at each generation step.
Grammar-constrained decoders mask tokens that leave the target language
\citep{scholak-etal-2021-picard,geng-etal-2023-grammar}, token-aligned
automata and generation engines make these checks exact and efficient for
subword models \citep{koo-etal-2024-automata,beurer-kellner-etal-2024-domino,
dong-etal-2025-xgrammar}, and type or semantic checkers extend the decoding
state beyond surface syntax
\citep{poesia-etal-2022-synchromesh,mundler-etal-2025-type}. These methods
keep every prefix extendable to some valid output and leave the length of
that extension open. Two lines of work reason about completion more directly.
ChopChop frames constrained decoding as the realizability of a prefix and
analyzes sound and complete monitors \citep{nagy-etal-2026-chopchop}.
Distance-guided decoding introduces the output budget: it precomputes token
distances to acceptance over a pushdown automaton with a bounded stack and
keeps a witness continuation whose first token always remains a candidate,
guaranteeing acceptance within the budget for context-free grammars
\citep{collura-etal-2026-bounds}. SBN well-formedness involves
non-context-free dependencies, and the cost of completing a prefix depends on
its full structural state. \method brings the witness idea to this setting:
it separates completion proposal from exact verification and keeps a verified, budget-feasible completion as a fallback, so every committed prefix retains a valid continuation within the remaining budget. Grammar-aligned
decoding and sequential Monte Carlo approximate the model distribution
conditioned on validity
\citep{park-etal-2024-grammar-aligned,loula-etal-2025-syntactic}; we study
greedy decoding, which returns one output per input.

\paragraph{Constraints after decoding.}
Post hoc methods select well-formed outputs from a set of candidates
\citep{liu-2024-soft}, and IterGen and DTV backtrack to an earlier point when
a check fails during generation
\citep{ugare-etal-2025-itergen,zhou-etal-2026-verifier}; unlike APD, both retract emitted tokens. Filtering is rejection sampling, whose cost on an input grows with that input's rate of invalid samples. Repair can account for much of what constrained decoding appears to gain: in tool-call abstention, the gains largely reflect the repair of unreadable outputs \citep{lee-2026-repair}. We compare filtering with constructive completion
under the same automaton and examine why filtering stays unreliable with
more candidates.

\paragraph{Constraints during training.}
Unlikelihood and margin losses on explicit negative tokens discourage
undesirable continuations
\citep{welleck-etal-2020-neural,noji-takamura-2020-analysis}, and grammar
masks and symbolic losses add structural constraints to the learning
objective \citep{english-etal-2025-grammar,bunel-etal-2018-leveraging,
ahmed-etal-2023-pseudo}. We ask when such supervision adds information
beyond likelihood training, and relate its effect to the probability mass the model leaves on rejected tokens.

\paragraph{DRS and semantic parsing.}
Neural DRS parsers commonly generate linearized meaning representations and
check well-formedness after generation
\citep{van-noord-etal-2018-exploring,fancellu-etal-2020-accurate,
wang-etal-2023-pretrained,zhang-etal-2024-gaining,zhang-etal-2025-retrieval,
zhang-etal-2025-neural}. DRS validity depends on references, scope,
accessibility, and graph structure as well as local syntax, which makes DRS
parsing a demanding testbed for combining neural generation with symbolic
constraints. Constrained decoding has shown particular benefits for semantic
parsing with little training data \citep{roy-etal-2023-benchclamp}.
We observe a similar dependence on
supervision and link each stage to a measurable property of the parser.

\section{One Automaton, Three Stages}
\label{sec:method}

\begin{figure*}[t]
\centering
\includegraphics[width=\textwidth]{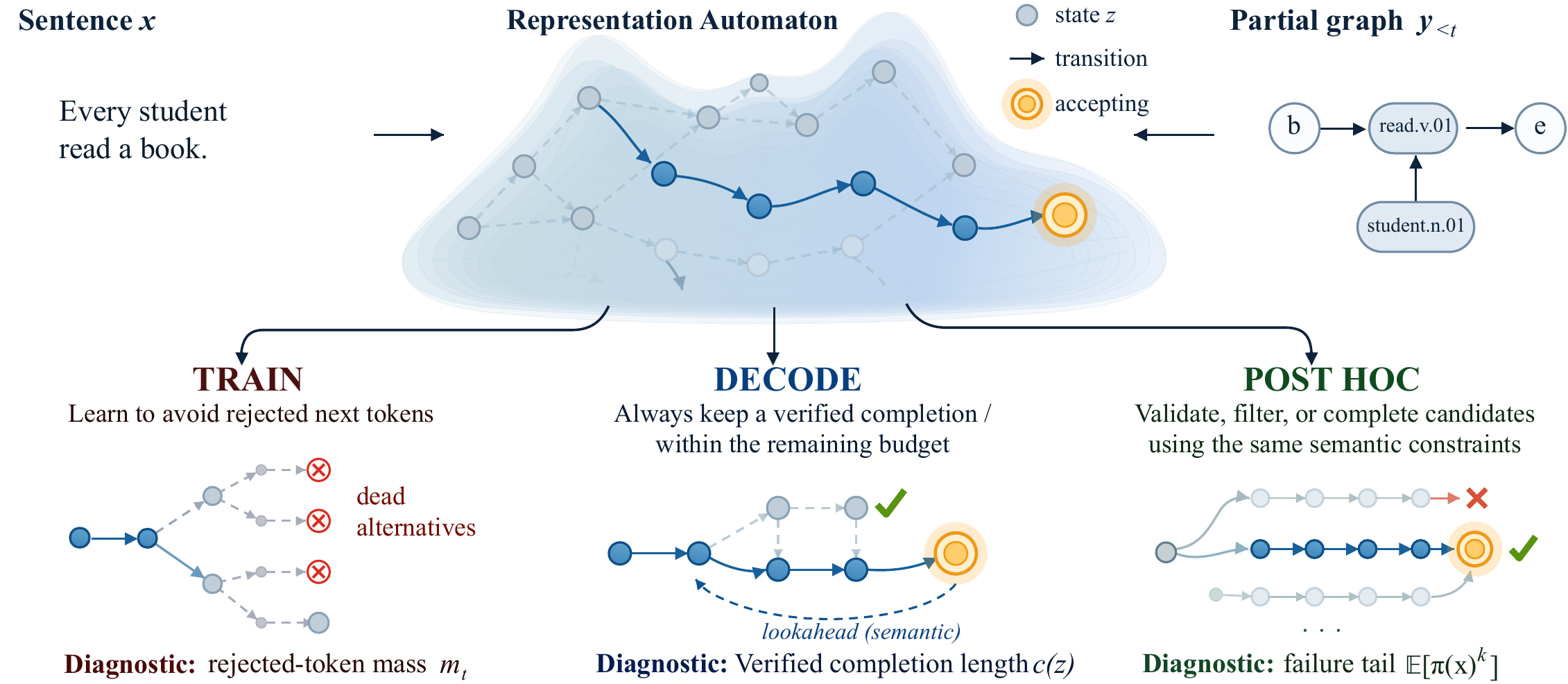}
\caption{A shared structural automaton across three stages of DRS parsing.
During training, it identifies rejected next tokens; during decoding, it keeps
a verified completion within the remaining output budget; after generation,
it validates, filters, or completes candidates. Each stage is characterized by
one quantity: the rejected-token mass $m_t$, the verified completion length
$|w|$ (an upper bound on the distance to acceptance), and the per-input
failure tail.}
\label{fig:overview}
\end{figure*}

Figure~\ref{fig:overview} shows how one structural automaton supports the
three stages of DRS parsing. We first define the distance to acceptance
(\S\ref{subsec:future}) and the SBN automaton (\S\ref{subsec:sbn-automaton}),
then describe decoding (\S\ref{subsec:decoding}), post hoc processing
(\S\ref{subsec:posthoc}), and training (\S\ref{subsec:training}).

\subsection{Distance to acceptance}
\label{subsec:future}

A neural parser $p_\theta(y\mid x)$ generates an output token by token from
its subword vocabulary (native tokens). A deterministic structural automaton
processes these tokens incrementally: the transition function $\delta(z,v)$
advances state $z$ on reading token $v$, or enters an absorbing rejection
state $\bot$. Accepting states $F$ mark complete, well-formed outputs ending in end-of-sequence (EOS).

We define the \emph{distance to acceptance} of a state $z$ as the minimum
number of additional tokens required to reach an accepting state:
\begin{equation}
    d(z)=\min\{|w|:\delta^*(z,w)\in F\},
    \label{eq:distance}
\end{equation}
where $d(z)=\infty$ if no accepting continuation exists. For example, a
forward reference such as \texttt{Patient +2} requires at least two more
concepts before the output can end. More generally, unresolved references
and open boxes impose requirements that must be met before acceptance. A
state is \emph{dead} if $d(z)=\infty$ and \emph{$b$-viable} if $d(z)\leq b$,
where $b$ is the remaining output budget. Local token masking prevents
transitions into $\bot$ but does not guarantee $b$-viability.

Rather than computing $d(z)$ exactly, we establish viability constructively.
A deterministic completion procedure, the \emph{proposer}, builds a suffix
$w$ from the requirements recorded in the current state, and exact replay
through the same automaton verifies it. If $\delta^*(z,w)\in F$ and
$|w|\leq b$, then $w$ is a \emph{certificate} of $b$-viability, with
$d(z)\leq|w|\leq b$. If no certificate is found, the candidate is treated as
uncertified, although it may still be viable. Appendix~\ref{app:repro}
details the proposer, the verification checks, and how the initial certificate is obtained.

\subsection{An SBN structural automaton}
\label{subsec:sbn-automaton}

We represent DRSs in Sequence Box Notation (SBN)
\citep{abzianidze-etal-2017-parallel,bos-2023-sequence}, which linearizes a
DRS into a sequence of concepts, roles, relative references, and box
operators. SBN well-formedness extends beyond local token syntax: references
must resolve to existing concepts, forward references must eventually be
resolved, box structure must support accessibility, and the typed graph must
be acyclic and, once complete, connected.

\begin{figure}[t]
\centering
\includegraphics[width=\columnwidth]{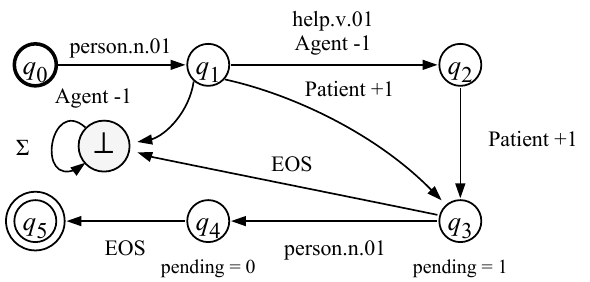}
\caption{Fragment of the SBN structural automaton. Forward references create
pending targets that must be resolved before acceptance. $\bot$ is the
absorbing rejection state and $\Sigma$ the token alphabet.}
\label{fig:sbn-automaton}
\end{figure}

Figure~\ref{fig:sbn-automaton} illustrates how the automaton handles these
dependencies. Once a concept has been introduced, a role may refer back to
it, whereas an out-of-range backward reference is rejected immediately. A
forward reference such as \texttt{Patient +1} instead creates a pending
target (\texttt{pending=1}). EOS is rejected in this state, and introducing a
later concept resolves the target and makes acceptance possible again. The
figure shows only a fragment; the full state also tracks box structure, the typed graph, and the accessibility relation.

The full state follows the checks of the PMB Referee
\citep{van-noord-etal-2018-evaluating} and has six components
(Table~\ref{tab:state} in Appendix~\ref{app:repro}): the lexical phase, the
concept register, pending targets, the box register, the typed graph, and
the accessibility relation. An exact byte-level interface maps the parser's
native tokens onto these symbols, including incomplete UTF-8 sequences, so
the automaton can evaluate every native token after any prefix. Unresolved
references, open boxes, and other requirements therefore remain explicit
throughout generation, and the proposer reads them to build completions.

\subsection{Decoding}
\label{subsec:decoding}

The local mask $W$
admits a token $v$ when $\delta(z,v)\neq\bot$, so every emitted token passes
the incremental checks and an emitted EOS passes full acceptance. A legal
prefix can still have a distance to acceptance exceeding the remaining
budget, and $W$ then runs out of tokens before the graph is complete.

A \emph{fixed reserve} $r$ stops $W$ after $H-r$ tokens and appends a
verified completion of at most $r$ tokens. Because the required length
depends on the state, a small reserve leaves requirements unresolved and a
large one stops the model early.

\emph{Viability-preserving decoding} commits to a token only when its
successor is certified $b$-viable. Its direct implementation, \emph{WF}
(fresh viability), certifies every candidate anew at each step, using the
proposer or a state--budget cache, and stops with a score of zero when no
candidate is certified. Its progress depends on fresh certificates at every step.

\emph{Accepting-path decoding} (\method) instead retains one verified
completion of its current prefix, its \emph{owned completion}, starting from
a certificate for the empty prefix. It leaves this completion only for a
higher-ranked token that is itself certified:

\smallskip
\noindent\begin{minipage}{\columnwidth}\small
\hrule
\smallskip
\textbf{\method{} step.}
State $z$, budget $b$, owned completion $h\cdot\bar w$.

\medskip
1.\ Examine the candidates $v$ that the model ranks above $h$, from the
highest. For the first $v$ whose successor $\delta(z,v)$ receives a
certificate $u$ within $b-1$ tokens, emit $v$ and own $u$.

\smallskip
2.\ If none is certified, emit $h$ and own $\bar w$.
\smallskip
\hrule
\end{minipage}
\smallskip

Each step either replaces the owned completion with another verified one or
advances along it, so every committed prefix keeps a verified completion
within the remaining budget. By induction, \method{} produces an accepted
output within the budget whenever the initial certificate exists, without
retracting any emitted token (Appendix~\ref{app:proofs}). An incomplete
proposer can keep \method{} from following a preferred token but cannot
invalidate the owned completion; WF has no such fallback.

\begin{proposition}[Lazy masking]
\label{prop:lazy}
With greedy decoding, identical model scores, and deterministic tie-breaking,
the local mask $W$ and unconstrained decoding $U$ agree on every token before
the first position $t^*$ at which a token from $U$ is rejected by $W$.
Regenerating the suffix from $t^*$ under $W$ therefore reproduces the masked
output, and $W=U$ whenever $W$ accepts every token of $U$.
\end{proposition}

\noindent
Outputs that never trigger a rejected transition thus need only validation,
and constrained regeneration starts only at the first rejected token
(Appendix~\ref{app:proofs}).

\subsection{Post hoc processing}
\label{subsec:posthoc}

Post hoc methods generate without constraints and consult the automaton
afterwards. \emph{Sample} draws eight outputs and returns the valid one with
the highest model probability. \emph{Beam} returns the first valid hypothesis
of a width-4 beam, a simplified form of \citet{liu-2024-soft}. Both filter
candidates and return an invalid output when every candidate is invalid.
\emph{Close} instead repairs: it backs off from the first rejected token to
the longest prefix with a certified completion within the budget and appends
that completion, replacing the model's content after that point
(Appendix~\ref{app:repro}).

\emph{W+Rollback} applies the same closure to the tokens generated by $W$:
it keeps the latest prefix $y_{1:t}$ with a certified suffix $w_t$ such that
$t+|w_t|\leq H$. It shares \method's proposer, verifier, budget, and initial
certificate, and retracts tokens whenever the latest certified prefix ends
before the generated output.

Filtering succeeds on an input when at least one candidate is valid. Let
$\pi(x)$ be the probability that a sampled output for input $x$ is invalid.
With $k$ samples drawn independently given $x$, the expected fraction of
inputs left without a valid output satisfies, by Jensen's inequality,
\begin{equation}
    \mathbb{E}_x[\pi(x)^k]\;\geq\;(\mathbb{E}_x\pi(x))^k,
    \label{eq:failure-tail}
\end{equation}
with equality for $k\geq2$ only when $\pi$ is constant across inputs. Mean
invalidity therefore understates the difficulty of filtering when failures
concentrate on some inputs.

\begin{proposition}[Filtering tail]
If $\pi(x)$ is beta-distributed with parameters $\alpha$ and $\beta$ across inputs, the expected
fraction of inputs with all $k$ independent samples invalid is
\[
    \frac{B(\alpha+k,\beta)}{B(\alpha,\beta)}
    = \Theta(k^{-\beta}),
\]
and the invalidity indicators of two samples for the same input have
correlation $\rho=1/(\alpha+\beta+1)$.
\end{proposition}

\noindent
Under this model, inputs with $\pi(x)$ close to one stay unresolved after
repeated sampling, which turns the exponential decay of independent failures
into polynomial decay. With mean failure rate $\bar p$,
$\beta=(1-\bar p)(1/\rho-1)$, so filtering depends both on how often the
parser fails and on how strongly its failures concentrate on the same
inputs.

\subsection{Training}
\label{subsec:training}

Cross-entropy identifies the correct next token but does not distinguish an
ordinary incorrect alternative from one the automaton rejects. At each gold
prefix with state $z_t$, we inspect the parser's $K=32$ most probable next
tokens and label a candidate \emph{rejected} when $\delta(z_t,v)=\bot$, using
the same transition test as $W$. Let $D_t$ be these rejected candidates; the
gold token always passes the test and is excluded. Their
\emph{rejected-token mass} is
\begin{equation}
    m_t=\sum_{c\in D_t}p_\theta(c\mid y_{<t}).
    \label{eq:dead-mass}
\end{equation}
Tokens outside the top $K$ are not penalized.

\emph{Unlikelihood} training (UL) adds
\begin{equation}
    \mathcal{L}_{\mathrm{ul}}
    =
    -\frac{1}{|T|}
    \sum_{t\in T}\log(1-m_t)
    \label{eq:ul}
\end{equation}
to the cross-entropy objective with weight one, where $T$ is the set of
target positions; the loss follows the aggregate form of
\citet{welleck-etal-2020-neural} and pushes probability away from rejected
alternatives. \emph{Masked training} instead removes $D_t$ from the softmax
before cross-entropy, approximating grammar-masked likelihood
\citep{english-etal-2025-grammar}. Masked candidates receive no gradient, so
the unmasked model is not taught to avoid them and inference still needs the
mask.

\begin{proposition}[Saturation]
\label{prop:saturation}
For a fixed candidate set $D_t$ and $m_t<1$, the gradient of
$-\log(1-m_t)$ with respect to the logits has $\ell_1$ norm $2m_t$.
\end{proposition}

\noindent
The derivative is $p_c$ for a rejected token $c$ and $-p_j m_t/(1-m_t)$ for
any other token $j$, so each group contributes $m_t$. Cross-entropy already
lowers the probability of every non-gold token, rejected ones included, so
UL reinforces an existing signal, and its added gradient vanishes as $m_t$
falls.

\section{Experimental Setup}
\label{sec:setup}

We vary two factors: the amount of gold supervision and the output budget.

\paragraph{Data.}
We use the gold English portion of PMB 5.1 in SBN format, with 9,057
training, 1,132 development, and 1,132 test graphs (sentences paired with
their DRSs). To vary supervision, we also train on random subsets of 600 and
1,256 training graphs. The automaton accepts every gold development and test
graph.

\paragraph{Models.}
We fine-tune Qwen3.5-0.8B-Base with rank-16
LoRA \citep{hu-etal-2022-lora} and select checkpoints by development loss. The decoding and post hoc experiments use one adapter per supervision level, and the training comparison trains its own CE and UL adapters.
Decoding is greedy, except that \emph{Beam} uses width 4 and \emph{Sample} draws eight outputs at temperature 1. Further details are in Appendix~\ref{app:repro}.

\paragraph{Output budgets.}
Budgets count native tokens, including EOS. The default budget $H=320$ is
about 1.3 times the longest gold output. The budget study uses $H=68$ and
$H=78$, the 90th and 95th percentiles of gold test length (68 and 79 on development), plus $H=58$, $100$, and $256$.

\paragraph{Evaluation.}
Validity is acceptance by the automaton, and semantic quality is
example-macro Smatch++ \citep{opitz-2023-smatch} with exact alignment,
scoring invalid outputs and decoder failures as zero. We also report a sensitivity score that zeros outputs rejected by the independent Referee check (Appendix~\ref{app:review-matrix}), counts of constructed placeholders, and paired bootstrap 95\% intervals.

\section{Results}
\label{sec:results}
\suppressfloats[t]

\subsection{Decoding}
\label{sec:budget}

\begin{table}[t]
\centering
\small
\renewcommand{\arraystretch}{1.12}
\caption{Same-budget test macro Smatch++ (graph validity \%).
Each arm has all 1,132 inputs. W and W+Rollback share one W trajectory decoded at $H=320$; WF and APD also require accepting EOS.}
\label{tab:budget}
\begin{tabular*}{\columnwidth}{@{\extracolsep{\fill}}rrrrr@{}}
\toprule
$H$ & W & WF & W+Roll. & \method \\
\midrule
68 & \shortstack[r]{86.27\\(93.5)} & \shortstack[r]{91.26\\(99.9)} & \shortstack[r]{91.27\\(100)} & \shortstack[r]{\textbf{91.33}\\(100)} \\
\addlinespace[3pt]
78 & \shortstack[r]{89.28\\(96.8)} & \shortstack[r]{91.61\\(99.8)} & \shortstack[r]{91.69\\(100)} & \shortstack[r]{\textbf{91.72}\\(100)} \\
\bottomrule
\end{tabular*}
\end{table}

Table~\ref{tab:budget} compares the decoders under the same budgets. APD
ends with an accepting EOS on every input and improves on W by +5.05 points
(95\% CI [3.94, 6.23]) at $H=68$ and +2.44 [1.67, 3.29] at $H=78$. Fixed
reserves close only part of this gap: in the original budget study
(Appendix~\ref{app:budget}), even the strongest
reserve, $r=16$, trails APD by 1.47 and 0.72 points and inserts placeholders
into 162 and 81 graphs. Table~\ref{tab:example} shows why: W is cut off
inside a concept, the reserve stops 16 tokens early and invents a \texttt{?}
target, and APD keeps the model's prefix and closes the open box.

\begin{table*}[t]
\centering
\small
\setlength{\tabcolsep}{4pt}
\caption{Example under the tight budget $H=58$. The local mask is truncated,
the fixed reserve closes early with a placeholder, and \method{} retains more
model-generated content while remaining within budget.}
\label{tab:example}
\begin{tabular}{@{}lp{12.5cm}@{}}
\toprule
\multicolumn{2}{@{}l}{\textit{Tom shook everybody's hands.}
\qquad budget $H=58$} \\
\midrule
Gold &
\texttt{male.n.02 Name "Tom" NEGATION <1 person.n.01 NEGATION <1
shake.v.09 Agent -2 Time +1 Patient +2 time.n.08 TPR now hand.n.01
PartOf -3} \\

$W$ &
\texttt{\ldots\ shake.v.01 Agent -2 Patient -1 Time +1 time.n.08 TPR now
CONTINUATION <0 person.n.01 hand.n.0}
\hfill \textit{cut off} \\

Reserve $r{=}16$ &
\texttt{\ldots\ shake.v.01 Agent -2 Patient -1 Time +1 time.n.08 TAB ?}
\hfill \textit{placeholder inserted} \\

\method{} &
\texttt{\ldots\ shake.v.01 Agent -2 Patient -1 Time +1 time.n.08 TPR now
CONTINUATION <1 person.n.01 EQU -3}
\hfill \textit{valid within budget} \\
\bottomrule
\end{tabular}
\end{table*}

\paragraph{Rescue and rollback.}
APD's gain over W comes from the 74 and 36 graphs that W truncates, which
contribute +5.12 and +2.47 points; on the graphs both decoders complete, the
score changes by $-$0.07 and $-$0.03 points. W+Rollback reaches nearly the
same scores (APD leads by +0.06 and +0.04 points, a tie once independent-checker rejections are also zeroed; Appendix~\ref{app:review-matrix}),
but it retracts 567 and 300 emitted tokens across 99 and 50 outputs (up to 23 and 27 per output) and repairs a W trajectory generated with $H=320$, so the two match in output length rather than in generation compute. APD keeps every token it emits. WF matches APD byte for byte on 1,131 and 1,130 inputs and stops on
the remaining one and two, where fresh search certifies no token.

\paragraph{Withheld proposals.}
We withhold new completions after initialization (Figure~\ref{fig:availability} in Appendix~\ref{app:review-matrix}). At withholding rates .25, .50, and .75, WF completes 799,
572, and 344 inputs, while APD completes all 1,132 and exceeds WF by 21.79,
25.04, and 27.62 Smatch++ points (every paired interval above zero, also for the sensitivity score).

\paragraph{Length and cost.}
Verified completions have a median length of four tokens and a 95th
percentile of 11, so most prefixes stay a few tokens from acceptance. APD follows its owned completion for 29\% of its decisions. 

\paragraph{AMR.}
We repeat the matched study on AMR~3.0 test graphs (1,898) at the 90th and 95th gold-length percentiles ($H=348$ and $411$; Appendix~\ref{app:amr}). APD
again gains +6.44 [5.55, 7.34] and +3.20 [2.56, 3.86] points over the local
mask, all from 183 and 93 rescued graphs, and ties WF and same-budget
rollback, which retracts 677 and 348 emitted tokens. Unlike on SBN, an eight-token reserve already yields a valid graph for every AMR input and comes within 0.06 points of APD.

\subsection{Post hoc processing}
\label{sec:posthoc}
\label{sec:supervision}

\begin{table}[t]
\centering
\small
\renewcommand{\arraystretch}{1.12}
\caption{Test Smatch++ at $H=320$ ($N=1{,}132$), with the invalidity of U
(\%) in parentheses. Sample/Beam leave 10.2/8.4\%, 2.5/3.2\%, and 0.4/1.6\%
of inputs invalid; development results are in Appendix~\ref{app:development}.}
\label{tab:supervision}
\begin{tabular*}{\columnwidth}{@{\extracolsep{\fill}}lrrr@{}}
\toprule
Method & 600 graphs & 1,256 & 9,057 \\
\midrule
U & 61.23 (23.9) & 73.19 (13.4) & 88.86 (4.4) \\
W & 75.24 & 81.29 & 91.88 \\
\method & \textbf{76.46} & \textbf{81.68} & \textbf{91.97} \\
\midrule
Close & 74.53 & 80.73 & 91.54 \\
Sample & 69.63 & 80.59 & 91.81 \\
Beam & 73.05 & 81.43 & 91.39 \\
\bottomrule
\end{tabular*}
\end{table}

Table~\ref{tab:supervision} compares decoding and post hoc methods at $H=320$ across three supervision levels. Unconstrained
generation fails on 23.9\% of test inputs with 600 training graphs and on
4.4\% with all 9,057. \method{} returns a valid graph for every input, and
its margin over the strongest post hoc method shrinks as supervision grows: +1.93 points (95\% CI [1.44, 2.44]) over Close with 600 graphs, +0.24 ([$-$0.55, 1.07]) over Beam with 1,256, and +0.16 ([$-$0.17, 0.50]) over Sample with 9,057, where both confidence intervals include zero.

\begin{figure}[t]
\centering
\includegraphics[width=\columnwidth]{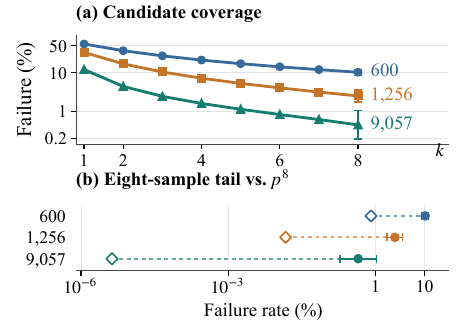}
\caption{Unresolved test inputs ($N=1{,}132$). Upper: failure among $k$
candidates, averaged over size-$k$ subsets of eight stored samples. Lower:
observed eight-sample tails (circles, Wilson 95\% intervals) versus $p^8$
(open diamonds), where $p$ is the mean sample failure probability. Colors
denote parsers trained on 600, 1,256, and 9,057 gold graphs.}
\label{fig:filter}
\end{figure}

\paragraph{Input clustering.}
At temperature one, a test sample is invalid with probability 55\%,
33\%, and 12\% at the three supervision levels, yet eight samples leave 10.2\%, 2.5\%, and 0.4\% of inputs without a valid candidate: 13, 168,
and more than $10^5$ times what independent failures at the same rate would
leave (Figure~\ref{fig:filter}). The within-input correlation stays at $\rho=0.27$, 0.26, and 0.28, and a beta-binomial fitted to the first four samples predicts the eight-sample tail
(Appendix~\ref{app:development}). The unresolved inputs are long and
persistent: with 600 graphs, 18.9\% of the longest quartile has no valid
sample, against 6.2\% of the shortest, and all five unresolved inputs at 9,057 graphs were already unresolved at 600.

\begin{table}[t]
\centering
\small
\setlength{\tabcolsep}{3pt}
\caption{First structural failures of unconstrained development outputs.
Only the lexical or syntactic row is visible to a context-free grammar.}
\label{tab:first-death}
\begin{tabular}{@{}lrrr@{}}
\toprule
First rejected token & 600 & 1,256 & 9,057 \\
\midrule
lexical or syntactic & 116 & 51 & 9 \\
forward reference unresolved & 46 & 21 & 13 \\
graph cycle & 45 & 25 & 2 \\
reference out of range & 39 & 28 & 8 \\
accessibility (Referee) & 11 & 2 & 3 \\
proposition box missing & 10 & 7 & 0 \\
\midrule
invalid outputs & 267 & 134 & 35 \\
\bottomrule
\end{tabular}
\end{table}

Non-context-free first failures become more common with supervision
(Table~\ref{tab:first-death}): with all 9,057 graphs they make up 74\% of
first failures on development and 64\% on test, beyond what a context-free grammar can check. They account for 37\% and 40\% of \method's surplus on invalid outputs at 600 and 1,256 graphs, below their 57\% and 62\% share of first failures, so APD's advantage tracks how often and how persistently a parser fails rather than the failure type.

\paragraph{Closing.}
Close repairs every invalid output by replacing content. At 600 graphs it trails \method{} by 1.93 points and puts placeholders in 133 development outputs, against 12 for \method{} (Appendix~\ref{app:development}).

\subsection{Training}
\label{sec:training}
\suppressfloats[t]

\begin{figure}[t]
\centering
\includegraphics[width=\columnwidth]{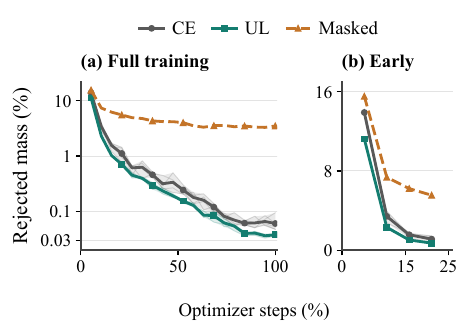}
\caption{Rejected-token mass at gold prefixes among top-32 tokens (600 training
graphs). Left: full training (log scale); right: first 25\% (linear).
CE/UL show five seeds (thin), means (thick), and min--max bands;
masked likelihood uses one seed.}
\label{fig:dead-mass}
\end{figure}

Figure~\ref{fig:dead-mass} tracks the rejected-token mass at gold prefixes. With 600 graphs, standard cross-entropy training (CE)
lowers it from about 14\% to below 2\% within the first sixth of training.
UL speeds this decline, ending near 0.04\%, against 0.05--0.09\% for CE
and 3--4\% for masked likelihood.

After 10\% of training, UL leads CE in development Smatch++ by 3.6, 11.0, and
3.9 points at the three supervision levels
(Figure~\ref{fig:learning-curves}). With 600 and 1,256 graphs, UL reaches the
validity and Smatch++ of the final CE checkpoint within 35--50\% of the
steps in all six runs; with all 9,057 graphs, it leads early and the two
curves meet at the end of training. Against development-selected CE checkpoints, three of the seven runs do so.

Final test effects vary across seeds (Appendix~\ref{app:training-endpoint}). At 600 graphs, UL changes unconstrained test
scores by +0.74 points on average over five seeds (range $-$4.15 to +5.01)
and scores under $W$ by $-$0.58; the larger settings give +0.87 and +0.33
unconstrained points. Automaton feedback thus mainly accelerates early
optimization: once cross-entropy has removed most rejected mass, decoding
constraints already enforce what it teaches.

\section{Discussion}
\label{sec:discussion}

\paragraph{Which stage to use.}
Structural constraints are most valuable when they provide information or
guarantees that model predictions alone cannot reliably supply. When the output
budget can bind, decoding needs a bounded check: local masking truncates
graphs, fixed reserves cut content, and APD completes every graph within the
budget; rollback reaches similar quality when emitted tokens may be revised
later. When the budget is generous, the per-input failure tail decides. With
full training data, few inputs lack a valid candidate: filtering comes within
0.2 points of \method{}, and lazy masking (Proposition~\ref{prop:lazy}) keeps
unconstrained generation batched and regenerates only the 4.4\% of outputs that
fail. With 600 training graphs, failures concentrate on persistent inputs: eight samples leave 10.2\% of inputs without a valid output, so constraints belong in decoding. Training-time feedback helps early, while
rejected-token mass is high, and complements constrained decoding: it shapes
the model at gold prefixes, whereas decoding governs the model's own prefixes.

\paragraph{Commitment and generality.}
Keeping every emitted token matters when outputs are consumed as they are generated, as in streaming or interactive parsing. APD adds little cost: it runs about as fast as WF (3,610 and 3,633 seconds over the test set at $H=68$), and both $W$ and APD take about 30--40\% longer per graph than unconstrained greedy decoding (Appendix~\ref{app:repro}). APD needs only an initial certificate, incremental transitions, and exact replay, and the filtering tail and rejected-token mass need only a validity checker, so each component carries over to other representations. What is specific to SBN is how strongly the distance to acceptance depends on the state. Closing a truncated AMR prefix takes at most eight tokens in our runs, mostly closing brackets, so a small fixed reserve suffices there; closing a truncated SBN prefix takes up to 40. Representations with long-range requirements, such as DRS, scoped meaning representations, or logical forms with variable binding, gain most from bounded decoding, because their distance to acceptance depends most on the state.

\section{Conclusion}
\label{sec:conclusion}

A single structural automaton serves DRS parsing at three stages, and each
stage is limited by a different quantity: decoding by the distance to
acceptance relative to the remaining budget, post hoc filtering by inputs
that are rarely parsed validly, and training feedback by the rejected-token
mass that cross-entropy quickly removes. Accepting-path decoding turns verified completions into a decoding guarantee: given an initial certificate, it reaches an accepted output within the budget without retracting tokens, and on DRS and AMR it completes budget-truncated graphs and matches same-budget rollback. Extending APD to sampling, and checking bounded viability lazily along unconstrained prefixes, are natural next steps. Where structural constraints should act thus depends on which structural failure remains.

\clearpage
\section*{Limitations}

Our three-stage analysis covers English DRS parsing in SBN with one model,
Qwen3.5-0.8B, fine-tuned with LoRA; AMR is evaluated for the decoding stage
only. Other meaning representations, languages, and larger or prompted
models remain to be tested in the same way.

All constrained decoders are greedy; lazy masking applies to greedy decoding
and, with shared noise, to sampling, but not to beam search. Supervision
levels come from random subsets of the gold data rather than naturally
low-resource corpora. Results with 600 graphs vary across seeds; training
end points use five seeds at this level and one run at each larger level,
and the decoding and post hoc analyses use one adapter per supervision
level. The beta-binomial model describes the observed failure tail within eight samples, is untested beyond them, and slightly underestimates it for the best parser.

Validity is acceptance by our automaton. Its lexical policy is broader than
the upstream CLF checker, which rejects some accepted outputs with
punctuation-initial lemmas or \texttt{x}-type senses
(Appendix~\ref{app:repro}), and incremental rejection is not a complete
coaccessibility test. Structural validity does not guarantee semantic
correctness, so we report semantic scores alongside validity.

\bibliography{custom}
\clearpage

\appendix
\raggedbottom
\setcounter{dbltopnumber}{3}
\renewcommand{\dbltopfraction}{.88}
\renewcommand{\dblfloatpagefraction}{.85}

\section{Matched Decoding Study}
\label{app:review-matrix}

\begin{figure}[t]
\centering
\includegraphics[width=\columnwidth]{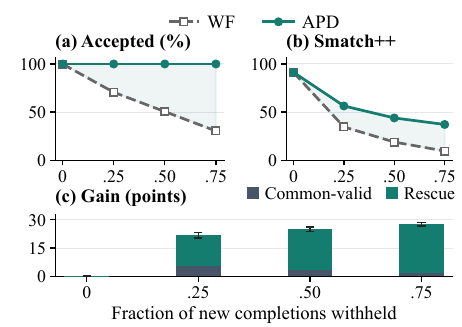}
\caption{Withholding new completions after initialization ($H=68$, test
$N=1{,}132$): (a) graph validity, (b) Smatch++, and (c) APD-minus-WF gain,
split into common-valid change and rescue, with paired 95\% intervals on the
total (Appendix~\ref{app:review-matrix}).}
\label{fig:availability}
\end{figure}

All methods in this study use the full-data adapter, all 1,132 test inputs,
and the same budgets, proposer, verifier, and withholding rule.
Table~\ref{tab:review-cells} lists every condition,
Table~\ref{tab:review-pairs} the paired differences,
Table~\ref{tab:review-decomposition} their decomposition,
Table~\ref{tab:review-operations} the proposer operations, and
Table~\ref{tab:review-certificates} the certificate statistics.

\begin{table*}[t]
\centering
\footnotesize
\renewcommand{\arraystretch}{1.0}
\setlength{\tabcolsep}{7pt}
\caption{All matched test conditions, $N=1{,}132$ per row. Wh.\ is the
fraction of new completions withheld after initialization. $L$ and $R$ count
valid outputs under the two validity policies, and $s_L$ and $s_R$ are the
corresponding Smatch++ scores. EOS counts outputs ending in an accepting EOS;
Fail counts decoder stops. W and W+Rollback share one W trajectory decoded
at $H=320$.}
\label{tab:review-cells}
\begin{tabular*}{\textwidth}{@{\extracolsep{\fill}}rrlrrrrrr@{}}
\toprule
$H$ & Wh.\ & Method & $L$ & $R$ & $s_L$ & $s_R$ & EOS & Fail \\
\midrule
68 & 0 & W & 1,058 & 1,058 & 86.27 & 86.27 & 1,033 & 0 \\
68 & 0 & W+Rollback & 1,132 & 1,131 & 91.27 & 91.20 & 1,132 & 0 \\
68 & 0 & WF & 1,131 & 1,129 & 91.26 & 91.12 & 1,131 & 1 \\
68 & 0 & APD & 1,132 & 1,130 & 91.33 & 91.19 & 1,132 & 0 \\
78 & 0 & W & 1,096 & 1,096 & 89.28 & 89.28 & 1,082 & 0 \\
78 & 0 & W+Rollback & 1,132 & 1,131 & 91.69 & 91.62 & 1,132 & 0 \\
78 & 0 & WF & 1,130 & 1,128 & 91.61 & 91.48 & 1,130 & 2 \\
78 & 0 & APD & 1,132 & 1,130 & 91.72 & 91.60 & 1,132 & 0 \\
256 & 0 & W & 1,130 & 1,130 & 91.88 & 91.88 & 1,130 & 0 \\
256 & 0 & W+Rollback & 1,132 & 1,131 & 91.97 & 91.90 & 1,132 & 0 \\
320 & 0 & W & 1,130 & 1,130 & 91.88 & 91.88 & 1,130 & 0 \\
320 & 0 & W+Rollback & 1,132 & 1,131 & 91.97 & 91.90 & 1,132 & 0 \\
\midrule
68 & .25 & WF & 799 & 741 & 34.59 & 32.28 & 799 & 333 \\
68 & .25 & APD & 1,132 & 1,071 & 56.38 & 53.67 & 1,132 & 0 \\
68 & .50 & WF & 572 & 501 & 18.88 & 16.50 & 572 & 560 \\
68 & .50 & APD & 1,132 & 967 & 43.92 & 37.76 & 1,132 & 0 \\
68 & .75 & WF & 344 & 291 & 9.53 & 8.04 & 344 & 788 \\
68 & .75 & APD & 1,132 & 924 & 37.16 & 30.46 & 1,132 & 0 \\
\bottomrule
\end{tabular*}
\end{table*}

\begin{table*}[t]
\centering
\footnotesize
\renewcommand{\arraystretch}{1.0}
\setlength{\tabcolsep}{7pt}
\caption{Paired Smatch++ differences, APD minus baseline, in points, over
all 1,132 inputs with paired bootstrap intervals. Exact counts byte-identical
outputs. $\Delta_L$ and $\Delta_R$ use validity policies $L$ and $R$.}
\label{tab:review-pairs}
\begin{tabular*}{\textwidth}{@{\extracolsep{\fill}}rrlrrr@{}}
\toprule
$H$ & Wh.\ & Baseline & Exact & $\Delta_L$ [95\% CI] & $\Delta_R$ [95\% CI] \\
\midrule
68 & 0 & W & 1,033 & 5.053 [3.938, 6.233] & 4.912 [3.790, 6.106] \\
68 & 0 & WF & 1,131 & 0.066 [0.000, 0.199] & 0.066 [0.000, 0.199] \\
68 & 0 & W+Rollback & 1,058 & 0.057 [0.016, 0.104] & $-$0.016 [$-$0.189, 0.092] \\
78 & 0 & W & 1,082 & 2.445 [1.672, 3.293] & 2.320 [1.568, 3.140] \\
78 & 0 & WF & 1,130 & 0.117 [0.000, 0.296] & 0.117 [0.000, 0.296] \\
78 & 0 & W+Rollback & 1,101 & 0.039 [0.015, 0.067] & $-$0.018 [$-$0.150, 0.060] \\
68 & .25 & WF & 59 & 21.788 [20.307, 23.281] & 21.396 [19.756, 23.053] \\
68 & .50 & WF & 5 & 25.044 [23.801, 26.281] & 21.259 [19.726, 22.744] \\
68 & .75 & WF & 0 & 27.624 [26.677, 28.559] & 22.424 [21.227, 23.619] \\
\bottomrule
\end{tabular*}
\end{table*}

\begin{table*}[t]
\centering
\footnotesize
\renewcommand{\arraystretch}{1.0}
\setlength{\tabcolsep}{7pt}
\caption{Decomposition of the APD-minus-baseline difference in Smatch++
points, each term divided by all 1,132 inputs. Counts list
common-valid/rescued/lost/neither inputs; the three contributions sum to the
total difference.}
\label{tab:review-decomposition}
\begin{tabular*}{\textwidth}{@{\extracolsep{\fill}}rrllrrrr@{}}
\toprule
$H$ & Wh.\ & Baseline & Policy & Counts & Common & Rescue & Lost \\
\midrule
68 & 0 & W & $L$ & 1058/74/0/0 & $-$0.065 & 5.119 & 0.000 \\
68 & 0 & W & $R$ & 1057/73/1/1 & $-$0.062 & 5.050 & $-$0.077 \\
68 & 0 & WF & $L$ & 1131/1/0/0 & 0.000 & 0.066 & 0.000 \\
68 & 0 & WF & $R$ & 1129/1/0/2 & 0.000 & 0.066 & 0.000 \\
68 & 0 & W+Rollback & $L$ & 1132/0/0/0 & 0.057 & 0.000 & 0.000 \\
68 & 0 & W+Rollback & $R$ & 1130/0/1/1 & 0.057 & 0.000 & $-$0.073 \\
78 & 0 & W & $L$ & 1096/36/0/0 & $-$0.026 & 2.471 & 0.000 \\
78 & 0 & W & $R$ & 1096/34/0/2 & $-$0.026 & 2.346 & 0.000 \\
78 & 0 & WF & $L$ & 1130/2/0/0 & 0.000 & 0.117 & 0.000 \\
78 & 0 & WF & $R$ & 1128/2/0/2 & 0.000 & 0.117 & 0.000 \\
78 & 0 & W+Rollback & $L$ & 1132/0/0/0 & 0.039 & 0.000 & 0.000 \\
78 & 0 & W+Rollback & $R$ & 1130/0/1/1 & 0.040 & 0.000 & $-$0.057 \\
68 & .25 & WF & $L$ & 799/333/0/0 & 5.479 & 16.309 & 0.000 \\
68 & .25 & WF & $R$ & 711/360/30/31 & 4.845 & 17.780 & $-$1.229 \\
68 & .50 & WF & $L$ & 572/560/0/0 & 3.439 & 21.606 & 0.000 \\
68 & .50 & WF & $R$ & 433/534/68/97 & 2.614 & 20.827 & $-$2.182 \\
68 & .75 & WF & $L$ & 344/788/0/0 & 1.932 & 25.692 & 0.000 \\
68 & .75 & WF & $R$ & 244/680/47/161 & 1.360 & 22.284 & $-$1.220 \\
\bottomrule
\end{tabular*}
\end{table*}

\begin{table*}[t]
\centering
\footnotesize
\renewcommand{\arraystretch}{1.0}
\setlength{\tabcolsep}{4pt}
\caption{Proposer operations of WF and APD, as totals over the test set,
including 1,132 initial constructor calls per row. Carry and Replan count
APD's token decisions along its owned completion and with a newly certified
completion. Times are in seconds, summed over inputs.}
\label{tab:review-operations}
\begin{tabular*}{\textwidth}{@{\extracolsep{\fill}}rrlrrrrrrrr@{}}
\toprule
$H$ & Wh.\ & Method & Requests & Withheld & Unknown & Construct & Carry & Replan & Proposal s & Gen. s \\
\midrule
68 & 0 & WF & 99,607 & 0 & 50,298 & 100,739 & -- & -- & 119.7 & 3,633.2 \\
68 & 0 & APD & 109,586 & 0 & 73,648 & 110,718 & 14,504 & 35,938 & 95.6 & 3,609.7 \\
78 & 0 & WF & 88,264 & 0 & 38,220 & 89,396 & -- & -- & 166.8 & 3,704.5 \\
78 & 0 & APD & 73,771 & 0 & 37,265 & 74,903 & 14,673 & 36,506 & 139.8 & 3,666.4 \\
68 & .25 & WF & 8,504,965 & 2,126,385 & 8,453,636 & 6,379,712 & -- & -- & 11,523.6 & 19,784.1 \\
68 & .25 & APD & 539,670 & 134,743 & 497,821 & 406,059 & 15,687 & 41,849 & 505.4 & 4,928.9 \\
68 & .50 & WF & 26,641,788 & 13,320,449 & 26,589,250 & 13,322,471 & -- & -- & 19,737.0 & 32,233.6 \\
68 & .50 & APD & 1,072,730 & 536,411 & 1,028,401 & 537,451 & 18,103 & 44,329 & 1,006.4 & 5,974.8 \\
68 & .75 & WF & 63,106,007 & 47,331,194 & 63,049,347 & 15,775,945 & -- & -- & 15,090.9 & 34,705.5 \\
68 & .75 & APD & 1,886,153 & 1,414,625 & 1,846,604 & 472,660 & 22,189 & 39,549 & 936.6 & 6,138.2 \\
\bottomrule
\end{tabular*}
\end{table*}

\begin{table*}[t]
\centering
\footnotesize
\renewcommand{\arraystretch}{1.0}
\setlength{\tabcolsep}{8pt}
\caption{APD certificates and decisions. Returned counts all verified
completions, including initial ones; lengths are in native tokens. Changed
counts decisions that differ from the model's top token; the last column
gives the median remaining budget for changed and unchanged decisions.}
\label{tab:review-certificates}
\begin{tabular*}{\textwidth}{@{\extracolsep{\fill}}rrrrrrrr@{}}
\toprule
$H$ & Wh.\ & Returned & Median & p95 & Max & Changed/decisions & Slack ch./unch. \\
\midrule
68 & 0 & 37,070 & 4 & 11 & 26 & 316/50,442 & 0/41 \\
78 & 0 & 37,638 & 4 & 11 & 27 & 195/51,179 & 2/51 \\
68 & .25 & 42,981 & 4 & 11 & 36 & 14,254/57,536 & 35/39 \\
68 & .50 & 45,461 & 5 & 14 & 41 & 27,510/62,432 & 34/36 \\
68 & .75 & 40,681 & 5 & 16 & 47 & 37,145/61,738 & 34/37 \\
\bottomrule
\end{tabular*}
\end{table*}

\paragraph{Validity policies.}
Policy $L$ counts outputs accepted by the automaton and scores invalid
outputs and decoder failures as zero. Policy $R$ additionally scores as zero
the outputs that the independent Referee type-and-structure check rejects
after CLF conversion. APD and successful rollback end with an accepting EOS
inside $H$.

\paragraph{Rollback.}
For each $H$, W+Rollback searches backward along the W trajectory decoded at
$H=320$ for the latest prefix whose certified completion fits $H$, using
APD's initial certificate, proposer, and verifier, with no further model
calls. Its scores are close to APD's (Table~\ref{tab:review-pairs}), and
APD's small margins under policy $L$ disappear under policy $R$. Rollback
discards 0.50, 0.27, 0.02, and 0.02 tokens on average at $H=68$, 78, 256,
and 320 (at most 23, 27, 13, and 12) and spends 5.9, 1.2, 0.4, and 0.4
seconds in total on repair.

\paragraph{Decomposition.}
Table~\ref{tab:review-decomposition} splits each difference into the
contributions of common-valid, rescued, and lost inputs, each divided by
$N=1{,}132$ (Appendix~\ref{app:budget}). Against W, the gain comes from
rescued outputs, with slightly negative common-valid changes; against WF, it
comes from the one and two inputs where WF stops. With withheld proposals,
the gain combines rescued inputs with improvements on inputs that both decoders complete.

\paragraph{Operations and certificates.}
Table~\ref{tab:review-operations} counts proposal requests, withheld
requests, unknown outcomes, constructor calls, and APD's carried and
replanned tokens; times are summed over inputs, and proposer time is part of
generation time. Table~\ref{tab:review-certificates} reports the lengths of
verified completions, which are upper bounds on $d(z)$, and the remaining
budget after each decision, grouped by whether APD's token differs from the
model's top token.

\paragraph{Independent checker.}
Across the 18 conditions, 628 outputs accepted by the automaton are rejected
by the independent check: 534 for lexical format, such as
punctuation-initial lemmas or \texttt{x}-type senses, 93 for other reasons,
and one for a conversion failure. Policy $R$ reflects these rejections.

\section{AMR Decoding Study}
\label{app:amr}

The AMR study follows the matched DRS design on the AMR~3.0 test set
(1,898 graphs). The parser is Qwen3.5-0.8B-Base with a rank-16 LoRA adapter
trained on the AMR~3.0 training set, and the automaton is an incremental
PENMAN state over the same byte-level bridge. Budgets are the 90th and 95th
percentiles of gold test length, $H=348$ and $411$ (339 and 396 on
development). One local-mask trajectory decoded at $H=448$ yields P at both
budgets, same-budget rollback, fixed reserves, and close-after, as for DRS;
WF and APD are decoded at each budget. Validity is PENMAN acceptance by the
automaton, and semantic quality is example-macro Smatch++ with exact ILP
alignment. Table~\ref{tab:amr} reports the results. Rollback retracts 677
and 348 tokens across 184 and 95 outputs (at most 7 per output), and
close-after exceeds the budget by at most 7 and 8 tokens. APD and WF produce
identical outputs on every input.

\begin{table}[t]
\centering
\small
\setlength{\tabcolsep}{4pt}
\caption{AMR test Smatch++ (validity \%), $N=1{,}898$. Close after may
exceed the budget.}
\label{tab:amr}
\begin{tabular}{@{}lrrrr@{}}
\toprule
 & \multicolumn{2}{c}{$H=348$} & \multicolumn{2}{c}{$H=411$} \\
\cmidrule(lr){2-3}\cmidrule(lr){4-5}
Method & F1 & V\% & F1 & V\% \\
\midrule
P (local mask) & 69.76 & 90.4 & 73.34 & 95.1 \\
$r{=}2$ & 70.82 & 91.9 & 73.89 & 96.0 \\
$r{=}4$ & 74.57 & 97.6 & 75.67 & 98.7 \\
$r{=}8$ & 76.15 & 100 & 76.52 & 100 \\
$r{=}16$ & 76.06 & 100 & 76.49 & 100 \\
P+Rollback & 76.20 & 100 & 76.54 & 100 \\
WF & 76.20 & 100 & 76.54 & 100 \\
\method & \textbf{76.20} & 100 & \textbf{76.54} & 100 \\
\midrule
Close after & 76.21 & 100 & 76.54 & 100 \\
\bottomrule
\end{tabular}
\end{table}

\begin{table}[t]
\centering
\small
\setlength{\tabcolsep}{4pt}
\caption{AMR test Smatch++ with new completions withheld after
initialization ($H=348$). Fail: WF stops.}
\label{tab:amr-withheld}
\begin{tabular}{@{}lrrrr@{}}
\toprule
Rate & WF & \method & $\Delta$ [95\% CI] & Fail \\
\midrule
.25 & 20.35 & 23.14 & 2.79 [2.10, 3.47] & 19 \\
.50 & 4.60 & 6.82 & 2.22 [1.76, 2.70] & 17 \\
.75 & 0.66 & 1.65 & 0.99 [0.71, 1.29] & 41 \\
\bottomrule
\end{tabular}
\end{table}

With new completions withheld (Table~\ref{tab:amr-withheld}), APD completes
every input and leads WF at every rate. Scores fall sharply for both
decoders: AMR outputs average about 200 tokens, and APD closes each graph
along its owned completion once replacements are withheld.

\section{Budget Study Details}
\label{app:budget}

\begin{figure}[t]
\centering
\includegraphics[width=\columnwidth]{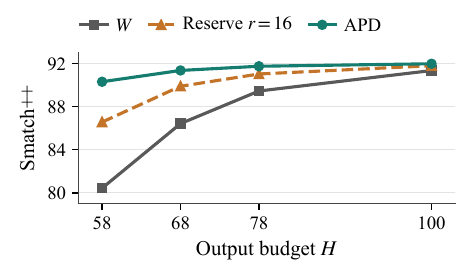}
\caption{Test validity and Smatch++ under output budget $H$ ($N=1{,}132$), from the original, independently decoded budget study (Appendix~\ref{app:budget}). The reserve stops W 16 tokens early.}
\label{fig:budget}
\end{figure}

This appendix reports the original budget study, in which each method is
decoded independently (Figure~\ref{fig:budget}, Table~\ref{tab:example}, and
Table~\ref{tab:reserves}). A fixed reserve $r$ stops the local decoder $r$
tokens before the budget and appends the constructed completion if it fits
within the reserved tokens, including EOS. \emph{Close after} appends the
same completion without enforcing the budget; it exceeds $H=58$, 68, 78, and
100 on 194, 99, 46, and 9 graphs by up to 40 tokens. Because these extra
tokens can raise its score, it serves as a reference rather than a
same-budget baseline. The local mask does not depend on $H$, so all reserve
and closing baselines reuse the local decoder's outputs without further model
calls.

\paragraph{Decomposition.}
Let $s_i^A$ and $s_i^B$ be the scores of \method{} and a baseline on input
$i$, with invalid outputs scored zero. We partition the inputs into
common-valid $C$, APD-only valid $R$, baseline-only valid $Q$, and
neither-valid inputs, and write the difference as
$\Delta=\Delta_{\mathrm{valid}}+\Delta_{\mathrm{rescue}}+\Delta_{\mathrm{lost}}$
with $\Delta_{\mathrm{valid}}=\frac{1}{N}\sum_{i\in C}(s_i^A-s_i^B)$,
$\Delta_{\mathrm{rescue}}=\frac{1}{N}\sum_{i\in R}s_i^A$, and
$\Delta_{\mathrm{lost}}=-\frac{1}{N}\sum_{i\in Q}s_i^B$.
Each term uses the full denominator $N=1{,}132$, and the intervals resample
paired inputs with the same draws for all terms. Here \method{} is valid on
every input, so $Q$ is empty.

\begin{table}[t]
\centering
\small
\setlength{\tabcolsep}{3pt}
\renewcommand{\arraystretch}{1.10}
\caption{APD-minus-baseline contributions to macro Smatch++ in points,
with paired 95\% intervals below. Each contribution is divided by the full
$N=1{,}132$, and the two terms sum to the total gain.}
\label{tab:budget-decomposition}
\begin{tabular}{@{}rlrr@{}}
\toprule
$H$ & Baseline & $\Delta_{\mathrm{valid}}$ & $\Delta_{\mathrm{rescue}}$ \\
\midrule
68 & $W$ &
\shortstack[r]{$-0.071$\\{\footnotesize$[-0.117,-0.030]$}} &
\shortstack[r]{$+5.009$\\{\footnotesize$[3.894,6.175]$}} \\
 & $r=16$ &
\shortstack[r]{$+1.307$\\{\footnotesize$[1.129,1.492]$}} &
\shortstack[r]{$+0.163$\\{\footnotesize$[0.000,0.380]$}} \\
\midrule
78 & $W$ &
\shortstack[r]{$-0.025$\\{\footnotesize$[-0.052,-0.001]$}} &
\shortstack[r]{$+2.331$\\{\footnotesize$[1.584,3.147]$}} \\
 & $r=16$ &
\shortstack[r]{$+0.555$\\{\footnotesize$[0.448,0.668]$}} &
\shortstack[r]{$+0.164$\\{\footnotesize$[0.000,0.378]$}} \\
\bottomrule
\end{tabular}
\end{table}

Table~\ref{tab:budget-decomposition} separates the two effects. Against $W$,
the gain comes from rescued outputs: of the 1,060 graphs that $W$ completes
at $H=68$, scores improve on 9, worsen on 17, and tie on 1,034 (5, 8, and
1,085 at $H=78$). Against $r=16$, 1,129 inputs are valid under both methods,
and most of the gain comes from this common-valid group, where \method{}
keeps model content that the reserve replaces.

\paragraph{Output length.}
On graphs that $W$ completes, \method{} and $W$ produce outputs of nearly
the same length (42.93 and 42.96 tokens at $H=68$; 44.14 and 44.17 at
$H=78$). On rescued inputs, $W$ exhausts the budget, while \method{}
averages 66.75 and 76.88 tokens and ends with an accepting EOS. At $H=58$,
one $W$ output ends in whitespace; we score its extracted SBN graph (0.79),
raising the score of $W$ by 0.07 points.

\begin{table*}[t]
\centering
\small
\renewcommand{\arraystretch}{1.12}
\setlength{\tabcolsep}{3pt}
\caption{Original test budget study. Each horizon has separate
columns for macro Smatch++ (F1), validity (V\%), and the number
of outputs containing constructed placeholders (PH). \emph{Close after} may exceed the budget.}
\label{tab:reserves}
\begin{tabular*}{\textwidth}{@{\extracolsep{\fill}}l*{12}{r}@{}}
\toprule
 & \multicolumn{3}{c}{$H=58$} & \multicolumn{3}{c}{$H=68$}
 & \multicolumn{3}{c}{$H=78$} & \multicolumn{3}{c}{$H=100$} \\
\cmidrule(lr){2-4}\cmidrule(lr){5-7}\cmidrule(lr){8-10}\cmidrule(lr){11-13}
Arm & F1 & V\% & PH & F1 & V\% & PH & F1 & V\% & PH & F1 & V\% & PH \\
\midrule
$W$ & 80.44 & 86.9 & 2 & 86.44 & 93.6 & 2 & 89.47 & 97.0 & 2 & 91.37 & 99.1 & 2 \\
$r{=}2$ & 80.84 & 87.8 & 20 & 86.22 & 93.5 & 11 & 89.30 & 96.7 & 4 & 91.43 & 99.2 & 3 \\
$r{=}4$ & 84.47 & 92.4 & 76 & 87.80 & 95.5 & 37 & 90.08 & 97.7 & 18 & 91.61 & 99.5 & 3 \\
$r{=}8$ & 85.91 & 95.0 & 149 & 89.10 & 97.5 & 74 & 90.51 & 98.5 & 42 & 91.62 & 99.5 & 7 \\
$r{=}16$ & 86.58 & 98.9 & 322 & 89.91 & 99.7 & 162 & 91.05 & 99.7 & 81 & 91.82 & 99.8 & 20 \\
\method{} & \textbf{90.33} & 100 & 3 & \textbf{91.38} & 100 & 2 & \textbf{91.77} & 100 & 2 & \textbf{92.00} & 100 & 2 \\
\midrule
Close after & 90.82 & 100 & 123 & 91.60 & 100 & 60 & 91.88 & 100 & 26 & 92.01 & 100 & 12 \\
\bottomrule
\end{tabular*}
\end{table*}

\section{Training End-point Details}
\label{app:training-endpoint}

\begin{figure}[t]
\centering
\includegraphics[width=\columnwidth]{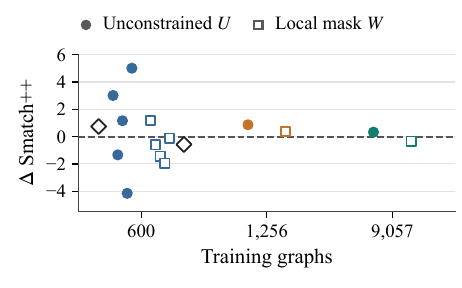}
\caption{Test Smatch++ change from CE to UL training ($N=1{,}132$), without
constraints (circles) and under the local mask (squares). Diamonds mark
five-seed means at 600 graphs; the larger settings have one run each.
Adapters are selected by development loss.}
\label{fig:training-test}
\end{figure}

\begin{figure*}[t]
\centering
\includegraphics[width=\textwidth]{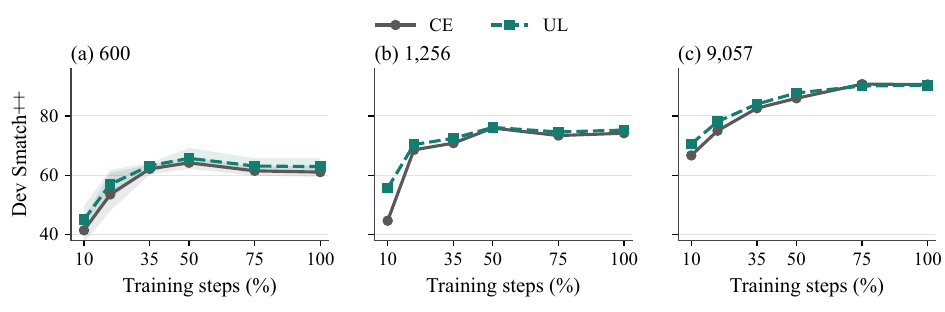}
\caption{Development learning curves for CE and UL training. Bands show the range over five seeds at 600 graphs.}
\label{fig:learning-curves}
\end{figure*}

Automaton-guided training changes the end point less consistently than early
training. With 600 graphs and adapters selected by development loss, UL
changes unconstrained development Smatch++ by +1.95 points on average over
five seeds (range $-2.39$ to +5.92; three seeds significantly positive and
one significantly negative) and by $-0.59$ under the local mask (one
significantly positive, three significantly negative). With 1,256 and 9,057
graphs, the unconstrained changes are $-0.15$ and $-0.27$, and the changes
under \method{} are +0.37 and $-0.16$, with intervals that include zero.
With 600 graphs, we ran \method{} on the seed with the largest unconstrained
gain: +1.28 points (95\% CI [0.71, 1.84]) on development and +0.83 on test.

On test, the five 600-graph seeds change by +0.74 points on average without
constraints (range $-4.15$ to +5.01) and by $-0.58$ under $W$; the
unconstrained changes for 1,256 and 9,057 graphs are +0.87 and +0.33, while
the corresponding development changes are negative.

Masked likelihood scores 3.23 without the mask at inference and 76.80 with
it, against 77.86 for standard training under $W$.

\section{Development Results and Failure Analysis}
\label{app:development}

The main comparisons use the test set; Table~\ref{tab:supervision-dev} gives
the corresponding development results.

\paragraph{Lazy masking on development.}
Across the three supervision levels, $W$ is identical to unconstrained
decoding whenever no unconstrained token is rejected (2,960 graphs). For 429
of the 436 outputs with a rejection, the two decoders also share the whole
prefix before the first rejected token; the remaining seven differ one token
earlier because of near-tied scores across separate runs. Only the 3--24\% of
outputs with a rejection therefore need constrained regeneration, and the
rest need only a validity check.

\begin{table}[t]
\centering
\small
\setlength{\tabcolsep}{2.5pt}
\caption{Development macro Smatch++ at $H=320$ ($N=1{,}132$).
Parentheses give U invalidity (\%).}
\label{tab:supervision-dev}
\renewcommand{\arraystretch}{1.12}
\begin{tabular*}{\columnwidth}{@{\extracolsep{\fill}}lrrr@{}}
\toprule
Method & 600 graphs & 1,256 & 9,057 \\
\midrule
U & 63.09 (23.6) & 76.10 (11.8) & 90.64 (3.1) \\
W & 77.86 & 83.68 & 92.56 \\
\method & \textbf{78.81} & \textbf{84.08} & \textbf{92.84} \\
\midrule
Close & 77.32 & 83.28 & 92.54 \\
Sample & 72.22 & 82.87 & 92.62 \\
Beam & 74.81 & 83.08 & 92.54 \\
\bottomrule
\end{tabular*}
\end{table}

\paragraph{Filtering tail.}
On development, a sample is invalid with probability 52\%, 31\%, and 11\%,
and eight samples leave 8.8\%, 2.6\%, and 0.6\% of inputs unresolved. A
beta-binomial fitted to the first four samples of each input predicts
eight-sample failure rates of 9.2\%, 2.9\%, and 0.41\%, close to the observed
8.8\%, 2.6\%, and 0.62\%, whereas independent failures predict 0.58\%,
0.009\%, and $2\cdot10^{-6}$\%. Fitted to all eight development samples, the
model predicts test failure rates of 9.29\%, 2.64\%, and 0.37\% (observed
10.16\%, 2.47\%, and 0.44\%). Table~\ref{tab:filter-uncertainty} gives Wilson
intervals for the observed test tail and input-bootstrap intervals for the
within-input correlation; the smallest tail has five inputs.

\begin{table}[t]
\centering
\small
\setlength{\tabcolsep}{4pt}
\caption{Test eight-sample tail and within-input correlation.
Tail intervals are Wilson; correlation intervals use an input bootstrap.}
\label{tab:filter-uncertainty}
\begin{tabular}{@{}lrrr@{}}
\toprule
Graphs & Unresolved & Tail \% [95\% CI] & $\rho$ [95\% CI] \\
\midrule
600 & 115 & 10.16 [8.53,12.06] & .273 [.250,.296] \\
1,256 & 28 & 2.47 [1.72,3.55] & .261 [.236,.285] \\
9,057 & 5 & 0.44 [0.19,1.03] & .278 [.238,.316] \\
\bottomrule
\end{tabular}
\end{table}

\paragraph{Structural failures on test.}
Replaying invalid unconstrained test outputs identifies forward-reference
failures, out-of-range references, missing proposition boxes, cycles, and
accessibility violations in 145/271, 88/152, and 32/50 first failures at the
three supervision levels. At full supervision, 64\% of first failures are
therefore invisible to a context-free grammar. First failures occur at
median positions of 71\%, 80\%, and 89\% of the output length, late in generation. 

\paragraph{Structural failures on development.}
Table~\ref{tab:first-death} in Section~\ref{sec:posthoc} gives the development failure audit.
Non-context-free failures account for 57\%, 62\%, and 74\% of first failures
at the three supervision levels. Close inserts constructed placeholders into
133, 61, and 20 development outputs, compared with 12, 8, and 1 for APD.

\section{Implementation}
\label{app:repro}

\subsection{Transitions and acceptance}
\label{app:checks}

Each native token is expanded to its bytes and replayed through a UTF-8
decoder and the SBN state; incomplete UTF-8 characters are buffered. An
invalid byte sequence, lexical prefix, or structural transition returns
$\bot$, and EOS additionally requires an empty UTF-8 buffer and a successful
final SBN check. Table~\ref{tab:state} lists the components. Cycle and
subordination-loop checks run as edges are added; unresolved
references, missing proposition boxes, and connectedness are checked at EOS.

\begin{table}[t]
\centering
\small
\setlength{\tabcolsep}{3pt}
\caption{State maintained by the SBN structural automaton. Lexical syntax
is local; other conditions depend on the evolving representation.}
\label{tab:state}
\begin{tabular}{@{}p{2.1cm}p{5.1cm}@{}}
\toprule
State component & Condition tracked \\
\midrule
lexical phase
& concept and role syntax;\newline constant and operator syntax \\
concept register
& backward targets already exist \\
pending targets
& forward targets are stored and must be resolved at EOS \\
box register
& references are range-checked; required proposition boxes must exist at EOS \\
typed graph
& cycles are rejected on insertion; connectedness is required at EOS \\
accessibility
& Referee-style box-subordination closure remains loop-free \\
\bottomrule
\end{tabular}
\end{table}

\paragraph{Lexical policy.}
A concept consists of a printable lowercase lemma without whitespace or
double quotes, a POS tag \texttt{n/v/a/r/x}, and a two-digit sense, so
constructed \texttt{.00} senses are accepted. The policy checks the surface
form of each concept; it admits punctuation- or hyphen-initial lemmas and
\texttt{x}-type senses, which the CLF Referee rejects.

\paragraph{References and boxes.}
A relative reference is converted to an absolute concept index. Backward
targets must exist, and forward targets are stored as pending edges;
introducing concept $i$ adds every pending edge to $i$ and checks the graph.
The typed graph contains box-to-concept containment, concept-role edges,
box-operator edges, and concept-to-box proposition edges, and an edge is
rejected when its target already reaches its source. Since a partial number
such as \texttt{-1} can still extend to \texttt{-10}, the check enumerates
existing targets whose index extends the current digits. Proposition targets
raise the number of required boxes, and \texttt{CONJUNCTION} makes boxes
aliases. Nesting, discourse edges, and cross-box concept uses add
subordination edges, closed under the Referee connectivity rule, and a
subordination loop is rejected. EOS requires a complete item, at least one
concept, a nonempty final box, no pending targets, all required boxes, and a
connected graph.

\paragraph{Independent check.}
We parse each raw SBN output in batch, without the incremental state,
convert it to CLF, and run the upstream PMB Referee. Its type-and-structure
check accepts all 2,264 gold development and test graphs; its check with the
official operator signature rejects 159 gold graphs because that inventory
lacks some operators, so we report it separately
(Table~\ref{tab:independent-audit}). Among the $H=320$ outputs, the
type-and-structure check rejects one WF and one APD output for the same
input, both with a hyphen-initial lemma.

\begin{table}[t]
\centering
\small
\setlength{\tabcolsep}{4pt}
\caption{Independent check on gold development and test graphs and on the
full-data test outputs at $H=320$. Counts are accepted graphs; Type/struct
and Signature are the two Referee checks.}
\label{tab:independent-audit}
\begin{tabular}{@{}lrrrr@{}}
\toprule
Graphs & $N$ & Automaton & Type/struct & Signature \\
\midrule
Gold & 2,264 & 2,264 & 2,264 & 2,105 \\
$U$ & 1,132 & 1,082 & 1,092 & 1,018 \\
$W$ & 1,132 & 1,130 & 1,130 & 1,052 \\
WF & 1,132 & 1,132 & 1,131 & 1,053 \\
APD & 1,132 & 1,132 & 1,131 & 1,053 \\
\bottomrule
\end{tabular}
\end{table}

\subsection{Completion proposer}
\label{app:constructor}

The proposer builds one completion from the parser phase and the requirements
recorded in the state, without model scores or gold information.

\paragraph{Concepts and pending targets.}
With an empty item and pending concept edges, the proposer adds
\texttt{entity.n.00} in the current box, repeating until every pending edge
is resolved, before opening required proposition boxes. An unfinished concept
is completed with a compatible POS tag and a \texttt{.00} sense; other
unfinished items are completed with the shortest compatible role or separator
string, breaking ties lexicographically.

\paragraph{Required boxes.}
While proposition boxes are still required, the proposer adds
\texttt{CONTINUATION <1} and a concept in the new box; pending concept edges
take priority. If a separator has started and a concept target is pending,
it finishes \texttt{CONJUNCTION} when the spelling allows, since box aliasing
preserves accessibility. A backward proposition target uses the nearest
existing box whose edge permits EOS, and otherwise the forward target
\texttt{>1}.

\paragraph{Roles, constants, and delimiters.}
At a trailing delimiter with nothing pending, the proposer adds \texttt{Of}.
An empty role target becomes \texttt{?} when that permits EOS, and otherwise
the nearest earlier concept that does. Open quoted strings are closed (with
\texttt{x} if empty), Name targets use \texttt{?}, and partial constants are
completed from the finite constant list or with zero digits, a denominator
of one, or \texttt{X} padding for quoted years. A complete reference or
constant is followed by a delimiter until EOS is possible.

\paragraph{Certification.}
The proposer rolls out fragments until the SBN state accepts EOS, stopping on
a repeated state, an empty proposal, a rejected fragment, or after
$\min(512,\max(32,8b))$ steps. It tokenizes the whole surface once, checks
that the tokens reproduce it byte for byte, appends EOS, and replays the
native-token suffix from $z$; the suffix becomes a certificate when it fits
$b$ and ends in $F$. When surface tokenization fails or $z$ holds incomplete
UTF-8, a fallback builds the suffix one fragment at a time. A failed
construction returns unknown, and \method{} replays every returned suffix
again before storing it.

\subsection{Initialization, repair, and training labels}

\paragraph{Initialization.}
From the empty state, the proposer suggests \texttt{entity.n.00} followed by
EOS, which becomes \method's initial certificate when it fits $H$. After
\texttt{person.n.01 help.v.01 Agent -1 Patient +1}, for example, adding
\texttt{entity.n.00} introduces concept 2 and resolves the pending Patient
edge, so this suffix plus EOS is a certificate whenever it fits the
remaining budget.

\paragraph{Close.}
Valid outputs are unchanged. For an invalid output, Close saves the states up
to the first rejected token and tries retained lengths $k$ from the longest
down to zero, calling the proposer with budget $H-k$; it returns the first
certified result, which exists whenever the empty-prefix certificate fits
$H$. If every attempt fails, the output is left unchanged.

\paragraph{Same-budget rollback.}
W+Rollback starts from the W trajectory decoded at $H=320$ and, for each
$H\in\{68,78,256,320\}$, keeps at most its first $H$ tokens. A trajectory
that already ends in an accepting EOS is returned unchanged; otherwise
retained lengths $t$ are tried from the longest down, with the proposer at
budget $H-t$, and the first certified $y_{1:t}w_t$ is returned, with \method's
empty-prefix certificate as fallback and no further model calls.

\paragraph{Training labels.}
Gold prefixes are replayed through the same transitions, and after a rejected gold prefix the structural loss is disabled. The implementation clamps $1-m_t$ at $10^{-6}$ before the logarithm, so Proposition~\ref{prop:saturation} holds wherever the clamp is inactive, with the candidate sets fixed during each backward pass of training.

\subsection{Training and runtime}

\paragraph{Hyperparameters.}
LoRA is applied to all language-model linear layers with rank 16, alpha 32,
and dropout 0.05; the learning rate is $10^{-4}$ and the effective batch size
is 16.

\paragraph{Withheld proposals.}
After initialization, the matched study withholds new completions at rates
0.25, 0.50, and 0.75 with $H=68$. A deterministic hash of the input, the
structural state, and the remaining budget decides each withholding, and WF
and APD share this rule. Initialization, cache hits, and
consumption of an owned completion are exempt from withholding. 

\paragraph{Runtime.}
Checking the top $K=32$ candidates at each target position adds about
0.26\,s per optimizer step with 600 training graphs and 0.92\,s with 9,057.
$W$ and \method{} take about 30--40\% longer per graph than unconstrained
greedy decoding, and Sample and Beam about 60--120\% longer. Code,
configurations, outputs, and scores will be released.

\section{Proofs}
\label{app:proofs}

\paragraph{\method{} guarantee.}
Let the budget count native tokens, including EOS, and suppose the initial
certificate $w_0$ satisfies $\delta^*(z_0,w_0)\in F$ and $|w_0|\leq b_0$.
The invariant is that the current state $z$ owns $w$ with
$\delta^*(z,w)\in F$ and $|w|\leq b$.

Writing $w=h\cdot\bar w$, emitting $h$ retains $\bar w$ at the successor
with budget $b-1$. Emitting an alternative $v$ installs a suffix verified
from $\delta(z,v)$ within $b-1$. Both actions preserve the invariant.
The owned head is always available while the state is not accepting, and each
step reduces $b$ by one. At $b=0$, the invariant requires an empty suffix and
therefore $z\in F$. Thus APD reaches an accepting state within the budget and
keeps every committed token. An unknown proposal only removes an alternative, so, given the initial certificate, the guarantee holds for any proposer.

\paragraph{Lazy masking.}
Let $t^*$ be the first token of greedy $U$ rejected by $W$. Before $t^*$,
the unconstrained argmax is admissible, so with identical scores and deterministic tie-breaking both decoders share $y_{<t^*}$ by induction, and regenerating from this prefix under $W$ repeats the masked computation. If $W$ accepts every token of $U$, then $U=W$.

\end{document}